\documentclass[sigconf]{acmart}

\usepackage{enumitem}
\AtBeginDocument{%
  \providecommand\BibTeX{{%
    \normalfont B\kern-0.5em{\scshape i\kern-0.25em b}\kern-0.8em\TeX}}}

\setcopyright{acmcopyright}
\copyrightyear{2021}
\acmYear{2021}
\acmDOI{xx.xxxx/xxxxxxxxx.xxxxxxx}

\acmConference[epiDAMIK 2021]{4th epiDAMIK ACM SIGKDD International Workshop on Epidemiology meets Data Mining and Knowledge Discovery}{Aug 15, 2021}{Virtual}
\acmBooktitle{epiDAMIK 2021: 4th epiDAMIK ACM SIGKDD International Workshop on Epidemiology meets Data Mining and Knowledge Discovery}
\acmPrice{15.00}
\acmISBN{978-1-xxxx-XXXX-X}

\begin{document}

\title{Modelling Major Disease Outbreaks in the 21\textsuperscript{st} Century: A Causal Approach}

\author{Aboli Marathe}
\email{aboli.rajan.marathe@gmail.com}
\affiliation{%
  \department{Dept. of Computer Engineering}
  \institution{Pune Institute of Computer Technology}
  \city{Pune}
  \country{India}
}

\author{Saloni Parekh}
\authornote{S. Parekh and H. Sakhrani assert joint second authorship.}
\email{saloniparekh1609@gmail.com}
\affiliation{%
    \department{Dept. of Information Technology}
   \institution{Pune Institute of Computer Technology}
  \city{Pune}
  \country{India}
}

\author{Harsh Sakhrani}
\authornotemark[1]
\email{harshsakhrani26@gmail.com}
\affiliation{%
  \department{Dept. of Information Technology}
  \institution{Pune Institute of Computer Technology}
  \city{Pune}
  \country{India}
}


\begin{abstract}

  Epidemiologists aiming to model the dynamics of global events face a significant challenge in identifying the factors linked with anomalies such as disease outbreaks. In this paper, we present a novel method for identifying the most important development sectors sensitive to disease outbreaks by using global development indicators as markers. We use statistical methods to assess the causative linkages between these indicators and disease outbreaks, as well as to find the most often ranked indicators. We used data imputation techniques in addition to statistical analysis to convert raw real-world data sets into meaningful data for causal inference.  The application of various algorithms for the detection of causal linkages between the indicators is the subject of this research. Despite the fact that disparities in governmental policies between countries account for differences in causal linkages, several indicators emerge as important determinants sensitive to disease outbreaks over the world in the 21\textsuperscript{st} Century.
\end{abstract}



\begin{teaserfigure}
 \centering
    \includegraphics[width=\textwidth, height=250pt]{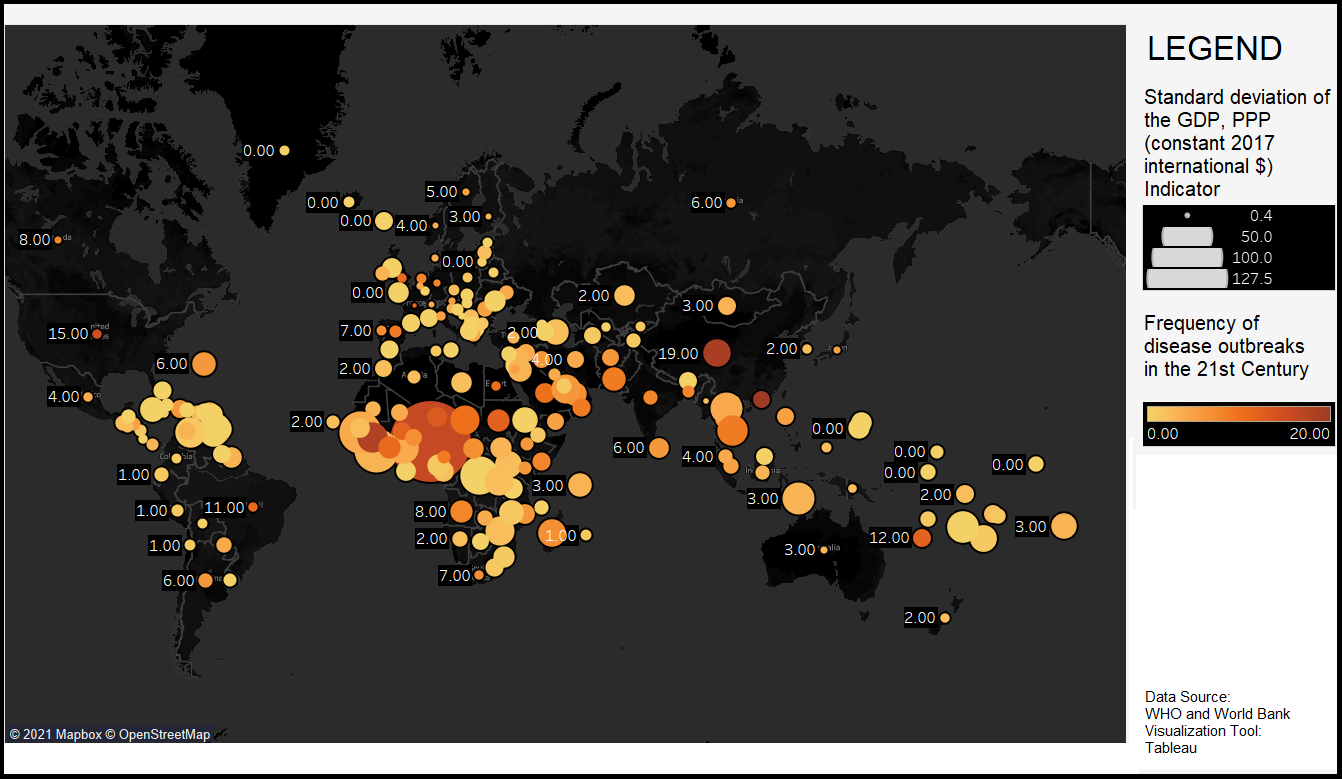}
    \caption{In the twenty-first century, the volatility in the GDP index seen on the globe appears to be linked to frequent disease outbreaks. We use statistical modelling to try to find causal links between similar indicators in this study.} 
    \label{fig1}
\end{teaserfigure}

\maketitle

\begin{figure*}[h]
 \centering
    \includegraphics[width=\textwidth, height=250pt]{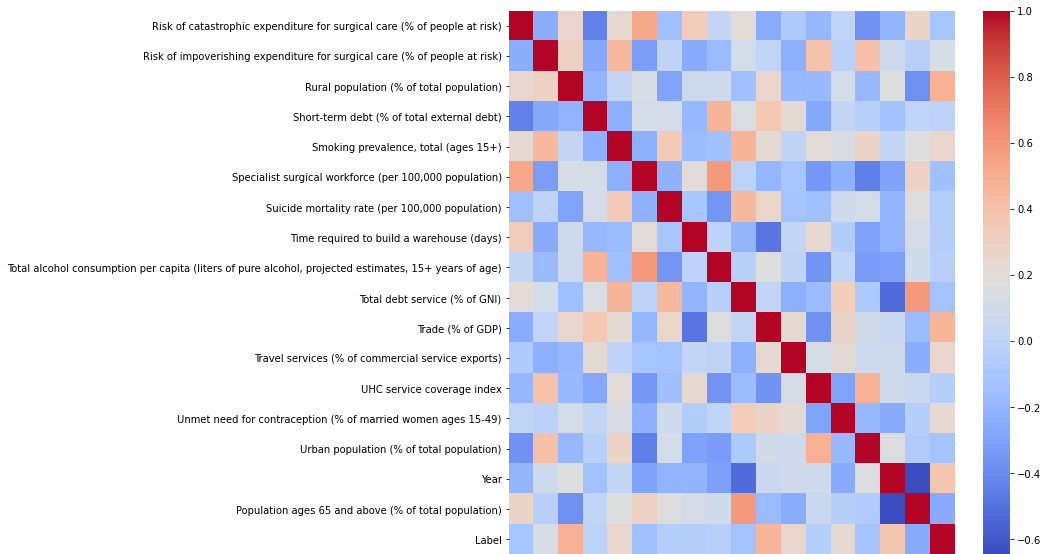}
\caption{Correlation heat map for indicators of St. Martin (French part). Few indicators are seen to be highly correlated with each other, and the disease outbreak occurrence has very weak correlations with other indicators.}
\label{fig2}
\end{figure*}

\section{Introduction}

Researchers have been studying the consequences of important events on global dynamics for centuries, with some focusing on socio-economic shifts, others on healthcare concerns, and yet others on cultural and historical factors. In the COVID-19 pandemic, the relationship between socio-economic factors and illness outbreaks has recently become a topic of great interest. Scientists around the world are still baffled as to how these outbreaks affect the planet or what elements influence them. These patterns are not only unpredictable, but they also vary from place to country due to differences in population, culture, geography, and other factors. For disaster management and outbreak preparedness, investigative analyses that lead to interpretable findings might be very beneficial.

We were inspired to perform this research after witnessing the terrible impacts of the COVID-19 pandemic. We wanted to learn more about the origins and effects of disease outbreaks around the world. We chose to approach the problem statement as a challenge in causal inference for this work, and we used statistical techniques to handle the data and derive conclusions from the indicator dataset. As a result, we use interpretable network diagrams to depict the features that have strong causal linkages to the incidence of disease outbreaks. The directionality between the nodes may show whether the outbreak was triggered by or impacted the preparedness in that sector. This study was carried out individually for each country after the missing data was imputed, and then the findings were aggregated for the entire world, following which the most commonly related nodes (indicators) were retrieved. These nodes indicate universal indicators linked to disease outbreaks, which authorities can analyze in depth in order to take necessary steps to assist the development sectors they represent.

\begin{table*}[h]
\caption{Sample of Granger Causality Matrix for St. Martin (French part)}
\begin{tabular}{llllllllll}\hline \hline
Feature &
  \begin{tabular}[c]{@{}l@{}}Electricity \\ Access\end{tabular} &
  \begin{tabular}[c]{@{}l@{}}National \\ Income\end{tabular} &
  \begin{tabular}[c]{@{}l@{}}Central \\ Govt. Debt\end{tabular} &
  \begin{tabular}[c]{@{}l@{}}Ext. Health \\ Expenditure\end{tabular} &
  GDP &
  Inflation &
  \begin{tabular}[c]{@{}l@{}}International \\ Tourism (Dept.)\end{tabular} &
  \begin{tabular}[c]{@{}l@{}}Mortality \\ (Diabetes, etc)\end{tabular} &
  \begin{tabular}[c]{@{}l@{}}Label \\ (Disease \\ Outbreaks)\end{tabular} \\ \hline 
\begin{tabular}[c]{@{}l@{}}Electricity \\ Access\end{tabular}            & 1      & 0.9823 & 1      & 1      & 1      & 0      & 0.7308 & 1      & 1 \\
\begin{tabular}[c]{@{}l@{}}National \\ Income\end{tabular}               & 1      & 1      & 1      & 1      & 1      & 1      & 1      & 1      & 1 \\
\begin{tabular}[c]{@{}l@{}}Central \\ Govt. Debt\end{tabular}            & 1      & 1      & 1      & 1      & 1      & 1      & 1      & 0.0184 & 1 \\
\begin{tabular}[c]{@{}l@{}}Ext. Health \\ Expenditure\end{tabular}       & 0.9948 & 1      & 0.9935 & 1      & 0.9935 & 0.9999 & 0.993  & 0.0186 & 1 \\
GDP                                                                      & 1      & 1      & 1      & 1      & 1      & 1      & 1      & 1      & 1 \\
Inflation                                                                & 0      & 1      & 1      & 0.9905 & 1      & 1      & 0.892  & 1      & 1 \\
\begin{tabular}[c]{@{}l@{}}International \\ Tourism (Dept.)\end{tabular} & 0.0777 & 1      & 1      & 0.984  & 1      & 0.4913 & 1      & 0.0002 & 1 \\
\begin{tabular}[c]{@{}l@{}}Mortality \\ (Diabetes, etc)\end{tabular}     & 1      & 1      & 0.9981 & 0.7306 & 1      & 1      & 0      & 1      & 1 \\
\begin{tabular}[c]{@{}l@{}}Label \\ (Disease \\ Outbreaks)\end{tabular}  & 1      & 1      & 0      & 1      & 1      & 0.9995 & 1      & 0      & 1
\\ \hline \hline
\end{tabular}
\label{tab1}
\end{table*}

\section{Related Work}

The world development indicators have been very popular among researchers trying to quantify or model the dynamics of global systems.  Using these indicators, scientists have been able to determine if  growth and development spur improvement in governance \cite{a}, links between population and resources \cite{b},  change in the development outcomes associated with the activities initiated by the MDGs \cite{c} and between financial development and economic growth \cite{d}. Some papers note their shortcomings in obtaining extensive local data, but were able to find distinctive  causal chains between the features. Specifically in  the field of healthcare, there have been many attempts to find the effects of disease burden \cite{e}, whether differences in microbial diversity can explain patterns of age-adjusted AD rates between countries \cite{f} and how spillovers of zoonotic infectious diseases into the human population will be impacted by global environmental stressors \cite{g}. The recent COVID-19 pandemic saw a rise in research work in this area, with many papers attempting to correlate the effectiveness of policies with the curve of the pandemic. From the dynamic causal modelling of COVID-19 \cite{h} to effects of non-pharmaceutical interventions \cite{i}, causal inference has been gaining preference for providing interpretable insights through scientific studies. Under the narrow field of disease outbreaks, some researchers have suggested measures for sustainable development \cite{j}, have forecasted economic trends \cite{k} or  have studied the historical trends \cite{l} and presented their views on planning for better preparedness. We observed that although these works are present at large, the task of analysing the causal relationships between socio-economic factors and disease outbreaks with our dataset has not been explored at a global scale and we present the results of such global network analyses in this work.

\begin{figure*}[h]
 \centering
    \includegraphics[width=\textwidth,height=580pt]{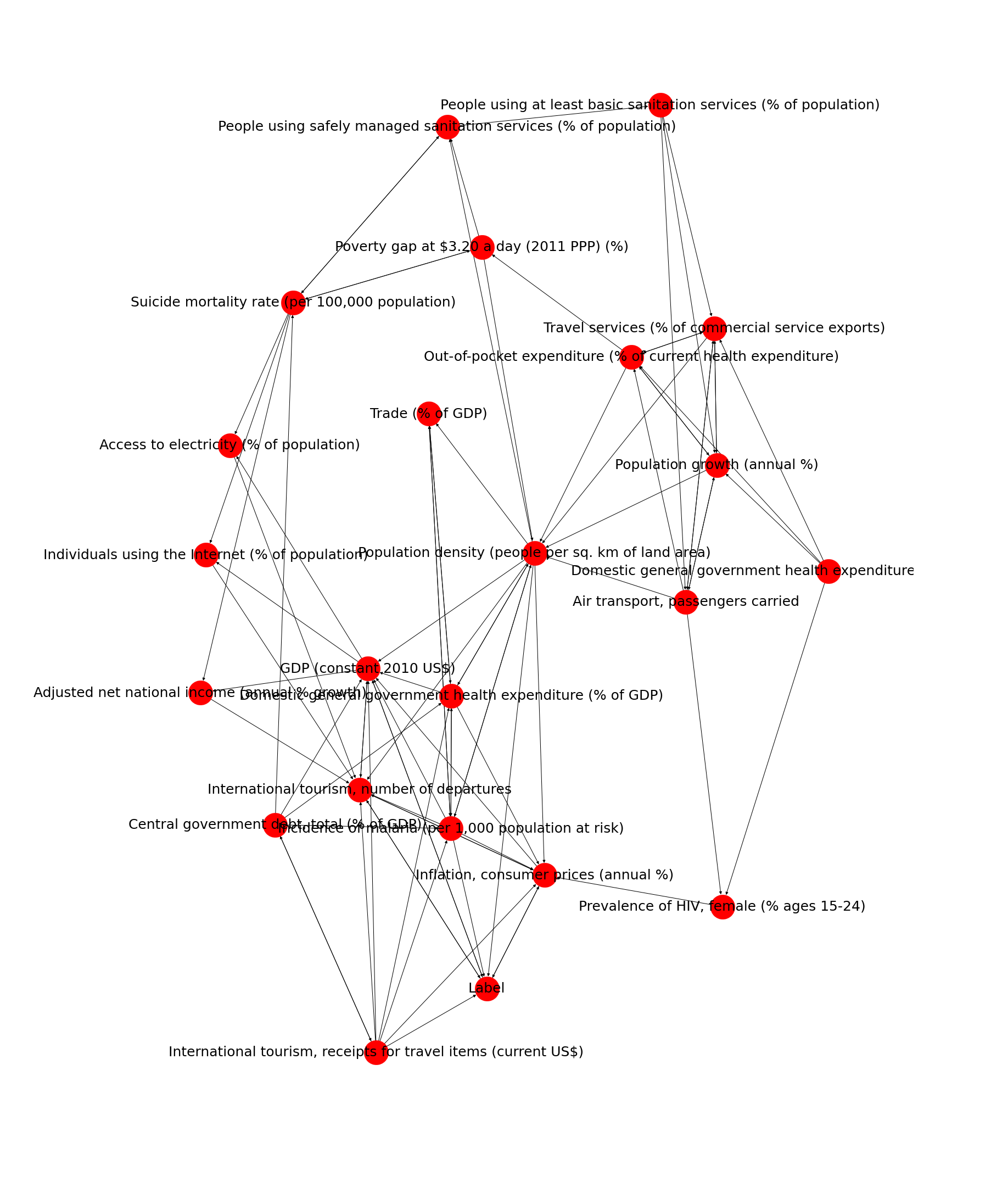}
\caption{Granger Causality Network for Bulgaria}
\label{fig3}
\end{figure*}

\begin{figure*}[h]
 \centering
    \includegraphics[width=\textwidth, height=580pt]{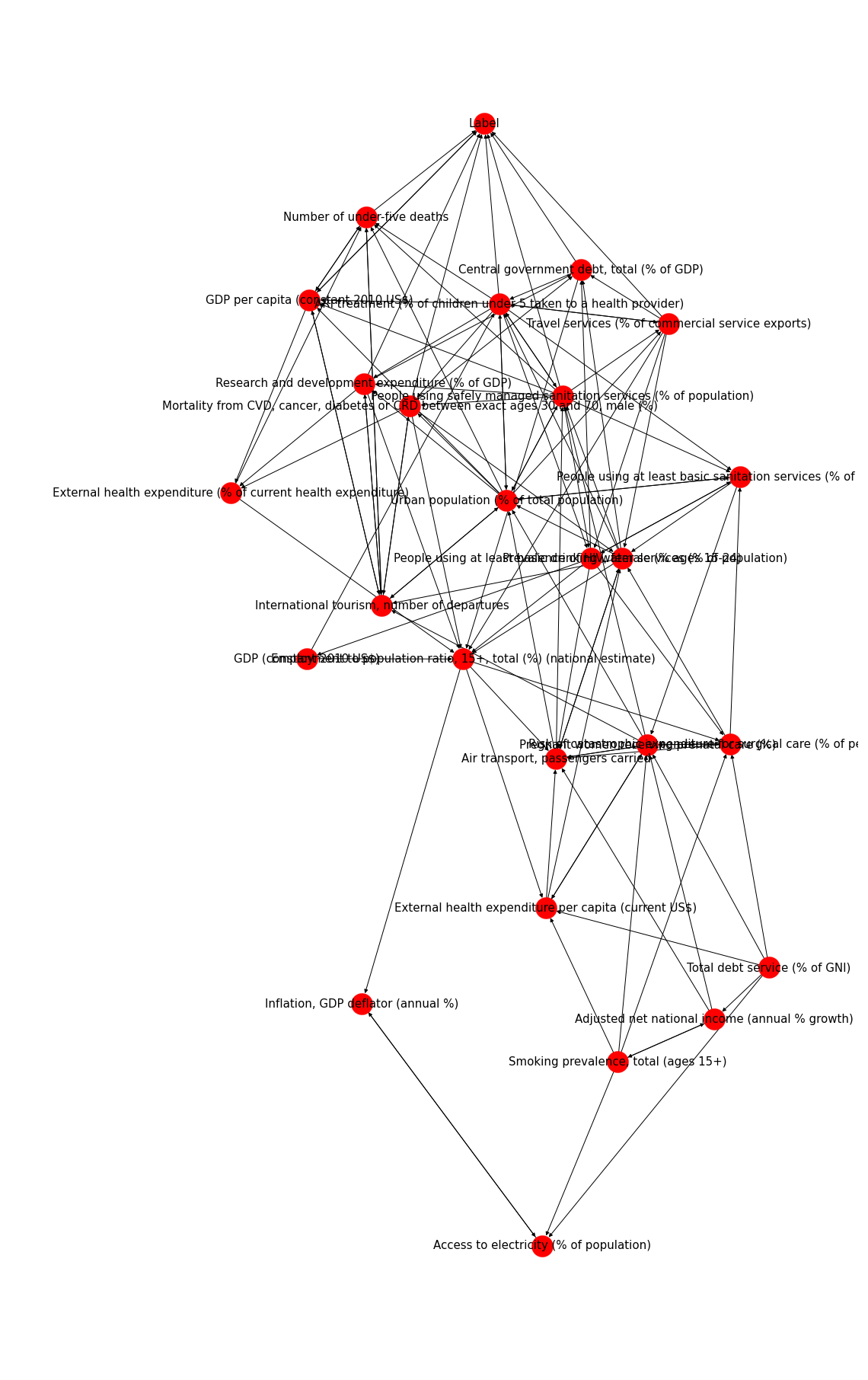}
\caption{Granger Causality Network for St. Martin (French part)}
\label{fig4}
\end{figure*}

\section{Methodology}

\subsection{Data Description and Creation}
The dataset used in this study was created from the World Development Indicators Data \cite{m} provided by the World Bank and the disease outbreak occurrence data by the World Health Organization(WHO) \cite{n}, put together to create a novel dataset for  determining the relationship between disease outbreak occurrence and socio-economic factors. World Development Indicators (WDI) is an expanding World Bank collection of development indicators from which we extracted 141 development indicators for 204 countries spanning over the years  2000 - 2019. Some examples of these indicators include ARI treatment (\% of children under 5 taken to a health provider) and Unmet need for contraception (\% of married women ages 15-49). The disease outbreak data from WHO was extracted separately for individual countries. The years that had an outbreak occurrence/absence were labelled as 1/0 respectively.

\subsection{Data Preprocessing and Statistical Tests}

The basic preprocessing involved encoding categorical features like country name, scaling the data and performing normalization. As the average percentage of missing values per column was 24\%, there was a need for data imputation techniques for filling the missing values. We employed a number of statistical data imputation techniques (KNN imputation, MSREG and Random Imputation) out of which MSREG provided the most relevant results for the analysis. We determined the effectiveness of the imputation algorithms by observing the statistical changes in the dataset before and after imputation, including variance, covariance and correlations.

The Stochastic Multiple Regression Imputation (MSREG) \cite{zz} method assigns values to each missing element $\dot{x}_{ir}$ according to (1), where $k$ is the number of manifest variables used in a model, $N_{m}$ is the number of missing values in $x_{i}$, and $Srandn()$ is a function that returns a different element of a standardized normally distributed random column vector each time it is invoked.
\begin{equation}
\dot{x}_{ir}  =  \sum_{j=1}^{k}  \hat{\beta}_{x_{i}x_{j}}x_{jr} + ( \sqrt{(1- \sum_{j=1}^{k} \hat{\beta}_{x_{i}x_{j}} \hat { \Sigma }_{x_{i}x_{j}}} )) \text{Srandn()}
\end{equation}
{\centering where $j = 1$ ... $k$,   $j \neq i$,   $r = 1$ ... $N_{m}$\par}

Some features had non-Gaussian distributions before and after imputation, thus changing them to exponential format transformed the dataset to a normal distribution. The Shapiro Wilk test \cite{o} (2) along with the histogram visualization was used to test the normality.  In this test, W  statistic tests whether a random sample, $x_{1}, x_{2} … , x_{n}$ comes from (specifically) a normal distribution. Small values of  W  are evidence of departure from normality and percentage points for the  W  statistic, obtained via Monte Carlo simulations.
\begin{equation}
W =  \frac {{ ( \sum_{i=1}^{n} a_i x_{(i)} ) } ^ 2 } {  \sum_{i=1}^{n}{( x_{i} - \overline{x} ) } ^ 2 } 
\end{equation}

where the $x_{(i)}$  are the ordered sample values  and the $a_i$  are constants generated from the means, variances and covariances of the order statistics of a sample of size n  from a normal distribution.
After performing the normality test, we tested if the data was stationary or not, as the format of the dataset is time series. For testing this, we used the augmented Dickey–Fuller test (ADF) statistic \cite{p,q} (3) which  tests the null hypothesis that a unit root is present in a time series sample.  Around 20 \% of the features were found to be non-stationary, which we made stationary by differencing the series twice and repeated the test again.
The unit root test is carried out under the null hypothesis  $\gamma = 0 $ against the alternative hypothesis of $\gamma < 0$. Once a value for the test statistic (\ref{eq2}) has been obtained, it may be compared to the Dickey–Fuller test's relevant critical value.
\begin{equation}
DF_ \tau  = \frac {\hat{ \gamma }}{ SE(\hat{ \gamma })}
\label{eq2}
\end{equation}
 If the calculated test statistic is less (more negative) than the critical value, then the null hypothesis of $\gamma = 0$ is rejected and no unit root is present and thus the series is stationary.

\begin{figure*}[h]
 \centering
    \includegraphics[width=\textwidth]{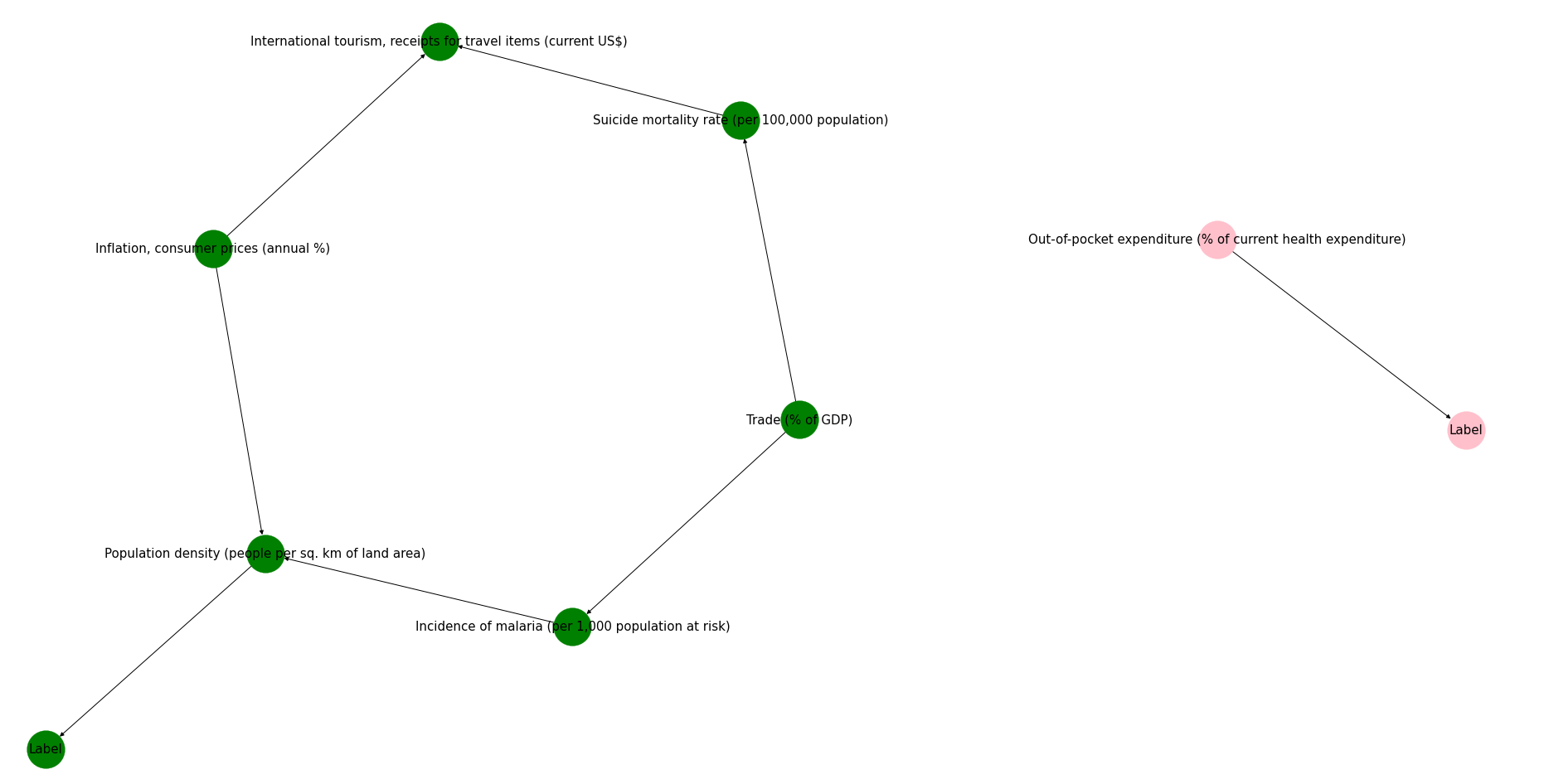}
\caption{IC* Algorithm Causality Network for Bulgaria}
\label{fig5}
\end{figure*}

\subsection{Learning Causal Relationships Between Indicators and Disease Outbreaks}
\leavevmode
\begin{enumerate}[label=(\alph*)]
\item Granger Causality Test 
\newline

Granger’s causality tests \cite{r,s,t}(4) the null hypothesis that the coefficients of past values in the regression equation is zero. This means the past values of time series (X) do not cause the other series (Y). So, if the p-value obtained from the test is lesser than the significance level of 0.05, then, we will  reject the null hypothesis. 
\begin{equation}
 \mathbb{P}[Y(t+1) \in A \vert \mathcal{I}(t)] \neq \mathbb{P}[Y(t+1) \in \vert \mathcal{I}_{-X}(t)] 
 \end{equation}

where  ${\displaystyle \mathbb {P}}$ refers to probability, ${\displaystyle {A}}$ is an arbitrary non-empty set, and ${\displaystyle {\mathcal {I}}(t)}$ and $ {\displaystyle {\mathcal {I}}_{-X}(t)}$ respectively denote the information available as of time ${\displaystyle t}$ in the entire universe, and that in the modified universe in which ${\displaystyle X}$ is excluded. If the above hypothesis is accepted, we say that ${\displaystyle X}$ Granger-causes ${\displaystyle Y}$.
\leavevmode\newline

\item IC* Algorithm
\newline

The IC* (Inductive Causation) algorithm \cite{u,v} can be used to
recover an underlying DAG structure from observed associations between traits. The algorithm is implemented as follows:
\begin{enumerate}
\item For each pair of variables a and b in ${V_O}$ search for a set $S_{ab}$ such that the  conditional independence between a and b given  $S_{ab}  ( a \bot  b| S_{ab})$ holds in  $ p({V_O})$. We begin by constructing an undirected graph linking the nodes a and b if and only if  $ S_{ab}$ is not found.
\item For each pair of non-adjacent nodes a and b with a common adjacent node c, we check if c belongs to  $ S_{ab}$
If it does, then continue and if not then we substitute the undirected edges by dashed arrows pointing at c.
\item Then we recursively apply the following rules:
\begin{itemize}
\item R1: For each pair of non-adjacent nodes a and b with a common neighbor c, if the link between a and c has an arrow head into c and if the link between c and b has no arrowhead into c, then add an arrow head on the link between c and b pointing at b and mark that link to obtain c –*-> b;  

\item R2: If a and b are adjacent and there is a directed path (composed strictly of marked links) from a to b, then add an arrowhead pointing toward b on the link between a and b;

\end{itemize}
\end{enumerate}

\end{enumerate}

\begin{table*}[h]
\caption{Frequency ranked indicators related to target variable}
\begin{tabular}{llllll}\hline \hline
\multicolumn{2}{l}{Granger Causality} &
  \multicolumn{2}{l}{\begin{tabular}[c]{@{}l@{}}IC* Algorithm\\ {[}Statistical Dependence{]}\end{tabular}} &
  \multicolumn{2}{l}{\begin{tabular}[c]{@{}l@{}}IC* Algorithm\\ {[}Genuine Causation{]}\end{tabular}} \\ \hline
Indicator &
  Freq. &
  Indicator &
  Freq. &
  Indicator &
  Freq. \\\hline
\begin{tabular}[c]{@{}l@{}}Individuals using the \\ Internet (\% of population)\end{tabular} &
  30 &
  \begin{tabular}[c]{@{}l@{}}Mortality rate attributed \\ to unsafe water\end{tabular} &
  14 &
  \begin{tabular}[c]{@{}l@{}}Out-of-pocket expenditure \\ (\% of current health expenditure)\end{tabular} &
  3 \\
\begin{tabular}[c]{@{}l@{}}GDP,  PPP (constant \\ 2017 international \$)\end{tabular} &
  28 &
  \begin{tabular}[c]{@{}l@{}}Central government debt,\\  total (\% of GDP)\end{tabular} &
  13 &
  \begin{tabular}[c]{@{}l@{}}Suicide mortality rate \\ (per 100,000 population)\end{tabular} &
  3 \\
\begin{tabular}[c]{@{}l@{}}GDP per person employed\\  (constant 2017 PPP \$)\end{tabular} &
  24 &
  \begin{tabular}[c]{@{}l@{}}People using safely managed \\ drinking water\end{tabular} &
  11 &
  \begin{tabular}[c]{@{}l@{}}Domestic general government\\  health expenditure\end{tabular} &
  3 \\
\begin{tabular}[c]{@{}l@{}}Inflation, consumer \\ prices (annual \%)\end{tabular} &
  24 &
  Trade (\% of GDP) &
  11 &
  \begin{tabular}[c]{@{}l@{}}Domestic general government\\  health expenditure\end{tabular} &
  1 \\
\begin{tabular}[c]{@{}l@{}}GDP \\ (constant 2010 US\$)\end{tabular} &
  24 &
  \begin{tabular}[c]{@{}l@{}}Individuals using the \\ Internet (\% of population)\end{tabular} &
  9 &
  \begin{tabular}[c]{@{}l@{}}People using at least \\ basic sanitation service\end{tabular} &
  1 \\
\hline \hline
  
\end{tabular}
\label{tab2}
\end{table*}

\begin{figure*}[h]
 \centering
    \includegraphics[width=\textwidth]{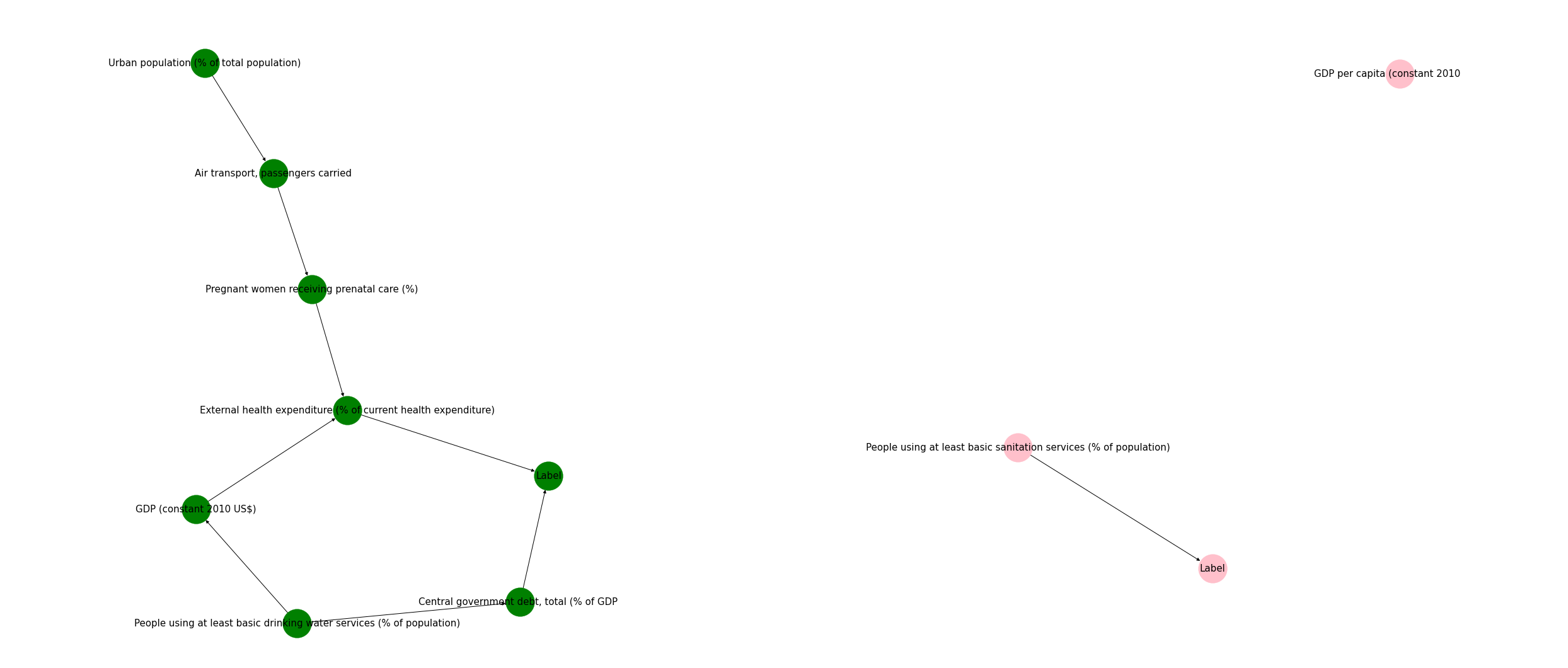}
\caption{IC* Algorithm Causality Network for St. Martin (French part)}
\label{fig6}
\end{figure*}

\begin{table}[]
\caption{Common indicators related to target variable identified by both Granger Causality and IC* algorithms}
\begin{tabular}{ll} \hline \hline
Country            & Indicator                                                                                      \\ \hline
Iran, Islamic Rep. & \begin{tabular}[c]{@{}l@{}}Hospital beds \\ (per 1,000 people)\end{tabular}                    \\
Iran, Islamic Rep. &
  \begin{tabular}[c]{@{}l@{}}People using at least \\ basic drinking water services\\ (\% of population)\end{tabular} \\
Liberia            & \begin{tabular}[c]{@{}l@{}}Incidence of malaria \\ (per 1,000 population at risk)\end{tabular} \\
Panama             & Trade (\% of GDP)                                                                              \\
South Sudan        & \begin{tabular}[c]{@{}l@{}}GDP per capita \\ (constant 2010 US\$)\end{tabular}                 \\
\begin{tabular}[c]{@{}l@{}}St. Martin \\ (French part)\end{tabular} &
  \begin{tabular}[c]{@{}l@{}}Central government \\ debt, total (\% of GDP)\end{tabular} \\ \hline \hline
\end{tabular}
\label{tab3}
\end{table}

\section{Analysis}

\subsection{Exploratory Data Analysis}
Before testing for causal relationships, we explored the data distribution, trends and characteristics of the 141 development indicators. To explore the data set, we calculated a correlation matrix using Pearson’s correlation coefficient \cite{w,x} and plotted the correlations in a heat map. A sample of this correlation heatmap for the country St. Martin (French part) can be seen in Figure \ref{fig2}. Some features were already heavily correlated and were removed to avoid erroneous connections in the final results. As data is stationary and fits normal distribution, it satisfies all the assumptions for the causality tests and we can proceed with the causal analysis.

\subsection{Granger Causal Analysis}

The first step was using the Granger causality values to construct a network showing predictive causal relationships between the nodes. We are trying to view only  the temporal relations through this statistic, as one thing preceding another can be used as a proof of causation. The Granger causality tests whether Y forecasts X, which could be interesting to observe in our indicator trends. The linkages were shown in the corresponding graphs. The total number of causal relationships between the target variable- occurrence of disease outbreaks and indicators was found to be 492 relationships.  Figures \ref{fig3} and \ref{fig4} show the Granger causality network graphs for Bulgaria and St. Martin (French portion) using causality matrices identical to the sample presented in Table \ref{tab1}.

\subsection{Application of IC* Algorithm}
By using this algorithm, we are essentially treating our problem statement as causal discovery with hidden variables and trying to remove irrelevant connections to maintain the potential causal connections thus inferring causal DAGs. Along with the algorithm, a Robust Regression Test \cite{y,z} was used to identify outliers and minimize their impact on the coefficient estimates. It also simultaneously checks the independence of the two time series. After applying this technique to each country separately, we observed several causal structures and their corresponding embedded patterns. The total number of causal relationships between the target variable- occurrence of disease outbreaks and indicators was found to be 234 relationships. In this graph, each variable is a node (green coloured nodes), and each edge represents statistical dependence between the nodes that cannot be eliminated by conditioning on the variables specified for the search. If the edge also satisfies the local criterion for genuine causation, then that network of directed edges has been isolated in graph 2 of each figure, marked by pink nodes. 11 such relationships of genuine causation were found in the dataset and are listed in Table \ref{tab2}. The IC* causality algorithm network graphs for Bulgaria and St. Martin (French part) are presented in figures \ref{fig5} and \ref{fig6}.

\section{Results}

After observing the graphs of 204 countries for 141 development indicators, we can clearly see that every country has a distinctive pattern of correlations and the total number causal relationships between features between the target variable- occurrence of disease outbreaks and indicators were found to be 492 relationships using the Granger Causality, 234 using IC* statistical dependence and 11 using the IC* genuine causation algorithm respectively. Out of the 234 relationships determined by IC*, only 6 were confirmed using both Granger and IC* algorithms which have been presented in Table \ref{tab3}. We observed the graphs obtained by the algorithms closely and noticed some interesting patterns. A certain subset of features were continuously found to be related with the target variable, the disease outbreak occurrence and have potential for genuine causation. By general observation, these features include indicators like individuals using internet, GDP, employment and health expenditure, which intuitively make sense as being factors affected by major disease outbreaks. By ranking these features by frequency, which can be observed in Table \ref{tab2}, the frequent features can be given to the authorities as preliminary findings, or can be fed to further network models to gain comprehensive insights. The main motivation behind this study was increasing the interpretability and attempting to trace the common causal relationships occurring in world dynamics over time which can be seen in the network graphs and ranked features. 

The findings provide easy-to-understand insights for the many nations included in the worldwide statistics.We can observe global patterns and country specific trends develop, and the direction of the impact seen in the directed graphs, provides us with insight on the nature of these connections, by aggregating the results gathered from all 204 nations. GDP and Healthcare Expenditure are some strong features that appear frequently in labelled outbreak sensitive features and can be targeted by authorities to become more resilient to the ravages of future outbreaks. 

\section{Limitations} 

One important observation is that the dataset in spite of containing over 140 indicators, is still not sensitive to the minor events and factors that influence modern countries. For example, the interactions between the employment ratio and pandemic occurrence may also be due to the ineffective policies or internal conflicts in the country. While critiquing the employed methodology,  we are aware that Granger causality is not necessarily true causality but can be indicative of the precedence of variables in the dataset. Directly utilizing the results of this study without a background verification for the given country may lead to incorrect assumptions about the nature of dynamics and further lead to ethical concerns by policy makers. Using the IC* algorithm to fine tune these results may potentially provide a degree of certainty to our determined causal relationships, but the verification of our results  using more complex causal algorithms may be necessary due to the complex nature of the data and randomness in world indicators. 

\section{Future Scope} 
This paper presents a new approach towards understanding how disease outbreaks affect development of countries across the world. In the future, we would like to extend this application, integrate more statistical analyses and build a more thorough knowledge framework based on the current dataset, combined with external country specific data sources. We would also like to share our insights with observations from domain experts studying the effects of disease outbreaks and provide better explanations for why each feature appears to have the respective causal relationship with the other features in a connected network.  The epidemiological findings may be utilised to build strong emergency preparation systems and plan and assess future development initiatives. We hope that this study will aid researchers in better understanding disease outbreak dynamics and their implications for global development. 

\small

\nocite{ph1,ph2,ph3,ph4,ph5,ph6,ph7,ph8,ph9,ph10,ph11,ph12}

\bibliographystyle{ACM-Reference-Format}

\end{document}